# Is plantar thermography a valid digital biomarker for characterising diabetic foot ulceration risk?

Akshay Jagadeesh[1], Chanchanok Aramrat[1], Aqsha Nur[1], Poppy Mallinson[1], Sanjay Kinra[1]

[1] Faculty of Epidemiology and Population Health, London School of Hygiene & Tropical Medicine, London WC1E 7HT, UK

## Abstract:

**Background** In the absence of prospective data on diabetic foot ulcers (DFU), cross-sectional associations with causal risk factors (e.g., peripheral neuropathy, and peripheral arterial disease (PAD)) could be used to establish the validity of plantar thermography for DFU risk stratification.

**Methods** First, we investigate the associations between the intrinsic clusters of plantar thermographic images with several DFU risk factors. We do this using an unsupervised deep-learning framework. We trained a Convolutional Neural Network Auto Encoder (ConvAE) to extract useful representations of feet-segmented thermographic images which we then hierarchically clustered. We then studied associations between obtained thermography clusters and DFU risk factors. Second, to identify those associations with predictive power, we used supervised learning to train Convolutional Neural Network (CNN) regression/classification models that predicted the risk factor based on the thermograph (and visual) input.

**Findings** Our dataset comprised 282 thermographs from type 2 diabetes mellitus patients (aged 56.31 ± 9.18 years, 51.42% males). On clustering, we found two overlapping clusters (silhouette score = 0.10, indicating weak separation). There was strong evidence for associations between assigned clusters and several factors related to diabetic foot ulceration such as peripheral neuropathy, PAD, number of diabetes complications, and composite DFU risk prediction scores such as Martins-Mendes, PODUS – 2020, and SIGN. However, models predicting said risk factors had poor performances (AUCs < 0.60, or large mean absolute errors).

**Interpretation** The strong associations between intrinsic thermography clusters and several DFU risk factors support the validity of using thermography for characterising DFU risk. However, obtained associations did not prove to be predictive, likely due to, spectrum bias, or because thermography and classical risk factors characterise incompletely overlapping portions of the DFU risk construct. Our findings highlight the challenges in standardising ground truths when defining novel digital biomarkers.

## Introduction

Up to a third of everyone with diabetes will develop a foot ulcer in their lifetime.[1] These ulcers do not heal very well, with nearly 60% of them getting infected, and 20% of those infected requiring surgical amputation.[1] Most amputations are preventable through the earlier identification and appropriate management of diabetic foot ulcers (DFU).[1] To this end, it is beneficial to reliably identify diabetes patients at high risk of developing DFU.[2,3] Screening such high-risk patients will facilitate the early detection of pre-ulcerative lesions and small DFUs and also help support risk factor modification (e.g., behavioural support for daily foot self-care, and offloading) and limb evaluation (vascular and/or nerve conduction studies). [1,2]

Studies have used infrared thermometer foot temperature measurements at specific foot sites to help risk stratify patients' DFU risks.[4] More recently, foot imaging with thermographic cameras (thermography) which have a higher spatial resolution (temperature estimates are provided across all points of the feet) and potential for automated/remote analysis,[5,6] is shown as a likely digital biomarker[7] for predicting DFU risk. Literature suggests thermographs may be able to pick up on pre-ulcerative (inflammatory) changes several weeks before visible skin changes, thereby allowing the identification of high-risk feet.[8–10] This modality could then offer an opportunity for regular screening by healthcare providers in settings where daily self-monitoring is unreliable. However, determining the validity of this approach will ideally require prospective on ulcers. Previous randomised controlled trials (RCTs) and prospective studies have provided inconsistent evidence that differing foot temperature distributions among diabetes participants are associated with unique DFU risks. A meta-analysis of five RCTs evaluating interventions of thermography coupled with personalised foot care and offloading for participants with hotspots (contralateral feet temperature differences ≥ 2.2 $^oC$) found some evidence for reduced DFU risk at any foot site (relative risk 0.51, 95% CI 0.31– 0.84).[10] However, several studies mostly included diabetes populations with previous ulcer histories.[10] Further, only one trial considered the outcome of foot ulcer development at, or adjacent to a previously identified hotspot, and they found no association.[10] We also argue that an RCT design may not be suitable for ascertaining the predictive capability of plantar thermographs in modelling DFU risk; the observed reductions in DFU risk could have occurred merely due to activity reduction even if it were done unrelated to any preceding thermographic changes. Only two previous prospective studies found thermographic findings could predict DFU with a lead time of 2 – 5 weeks.[8,9] However, one of them demonstrated an impractically high false positive rate of 57%,[8] and the other included only a very small sample (N= 30 patients) of high-risk diabetes patients with histories of previously healed ulcers.[9] Without prospective data, it is not possible to directly validate the use of thermography for predicting (absolute) DFU risk. Indirectly, however, we can use cross-sectional data to investigate whether plantar thermography is associated with classical DFU risk factors. Few previous studies have investigated such (predictive) associations.[11] Characterising risk factor severity, especially key ones like neuropathy and peripheral arterial disease (PAD) can then be used for risk stratification. Such prediction models would likely have additional clinical benefit; accurate classification into neuropathic and/or ischaemic foot will help tailor limb evaluation. Robust prediction model performances would then support the validity of using thermography as a digital biomarker for predicting DFU risk in diabetes populations.

In this study, we aimed to study the validity of plantar thermography as a digital biomarker to identify patients at high risk of diabetic foot ulceration (DFU). We limited ourselves to the plantar aspect as the literature shows most DFUs develop on the weight-bearing portions of the plantar aspect of the foot.[2] First, we investigate the associations between the intrinsic clusters of plantar thermographic images with DFU risk factors. We do this using an unsupervised deep-learning framework. We trained a Convolutional Neural Network Auto Encoder (ConvAE)[12,13] model to extract useful representations of feet-segmented thermographic images. We then applied hierarchical clustering to this dataset of thermography representations and explored associations between thermography clusters and DFU risk factors. Second, to identify those associations with predictive power we trained supervised learning-based Convolutional Neural Network (CNN) regression/classification models[14] that predicted the risk factor based on the thermograph (and visual) input.

## Methods

### Study population

The details of the APCAPS population have been described elsewhere.[15] Briefly, it is a prospective, inter-generational cohort based in the 29 villages of Ranga Reddy district in the South Indian state of Telangana. For this analysis, we used a subset of diabetes patients with available thermography images from the latest follow-up in 2022–2023.[16]

### Image acquisition and preprocessing

Thermal and corresponding registered visual images of the foot were acquired using the CAT S61 smartphone using a (non-insulating) cloth to cover non-foot portions according to the APCAPs data acquisition protocol.[17] The regions of interest demarcating the two feet in the thermal image were segmented using a CNN with U-Net architecture, models previously shown to have good segmentation performance in biomedical image processing.[18] Two authors CA, and AN created a ground truth dataset using ITK-Snap (version 4.0.2)[19] by independently drawing foot contours on a random sample of 60 visual images. We created a consensus foot segmentation mask from these two masks using the STAPLE (Simultaneous Truth and Performance Level Estimation) algorithm.[20] Then we trained the U-Net model to segment bilateral feet from visual images achieving high performance, intersection over union (IoU) score of 0.97. This trained segmentation model was used to obtain binary mask predictions from all visual images, and by applying these mask predictions to the corresponding registered thermal images, we extracted the thermal images of the feet for downstream analysis. Similar to previous studies, our segmentation model approach used registered visual images instead of direct segmentation on thermal images, as the morphological boundaries of the feet are not well demarcated in thermal compared to visual images.[6,21]

### Extracting image representations

We built a custom ConvAE model[12,13] consisting of an encoder with three layers of 2D CNN layers used to extract thermographic image representations that are dimensionally smaller than the original segmented thermal image. The decoder consisted of transposed layers trained to reconstruct output similar to the original data using representations generated from the encoder. The ConvAE model demonstrated good reconstruction (hold out set, mean squared error: 0.002; mean absolute error:

0.01). We then used this trained ConvAE model to extract representations (useful features) from all foot-segmented thermographs.

## Hierarchical clustering

We used the 'Ward' method of hierarchical clustering[12] to generate clusters within thermographic representations. We identified the optimal number of clusters for the dataset using the 'elbow' method[12] and visualised clustering results using t-SNE plots.[13] We report silhouette scores to quantify the quality of clustering.[22] We studied associations between DFU risk factors and thermographic cluster assignment using appropriate statistical tests.

## Prediction models

For DFU risk factors that showed strong associations with intrinsic thermography clusters, we supervised-learning-based trained several pre-trained and custom CNN classification/regression models[14] to predict those risk factors based on thermography (with visual image supplementation). Theoretically, it was thought that visual images would provide information regarding callosities, macerations between the toes, and foot deformities (including hammer or claw toes) that may aid in characterising DFU risk.[2] As a sensitivity analysis, we trained models to predict the thermography cluster based on thermography input.[23] Deep U-Net, ConvAE, and CNN prediction models were all trained using the Adam Optimiser with optimal learning rates identified through learning rate schedulers and stopped training when validation set performance metrics failed to show improvement in 10 over epochs.

## Results

### Associations between thermographic images and DFU risk factors

Our study included 282 participants (aged 56.31 ± 9.18 years, 51.42% males) with type 2 diabetes mellitus. Participants had diabetes for 5 (3 – 10) years with the majority never reporting having experienced an ulcer or amputation, 272 (96.45%). ConvAE obtained latent space representation for a sample thermograph shown in Figure 1. Hierarchical clustering was performed to visualise the intrinsic clusters within thermographs (Figure 2). In Panel A, the dendrogram shows two primary clusters with several sub-clusters demonstrating the distinct clusters of the ConvAE-generated representations of plantar thermographic images. In Panel B, the "elbow" point indicates the optimal number of clusters to be two. At two clusters the explained variance of the clustering begins to level off and the marginal gain of adding additional clusters decreases significantly. In Panel C, we visualise through t-SNE the distribution of the two distinguishable yet overlapping clusters of thermographic representations. In Panel D, we show that ConvAE has successfully captured hidden patterns of overlapping thermographic phenotypes finding visually discernible groups. However, the silhouette score of +0.10 suggests that the obtained clusters are weakly separated, indicating that while some distinct patterns are captured, there is still considerable statistical similarity between the clusters.

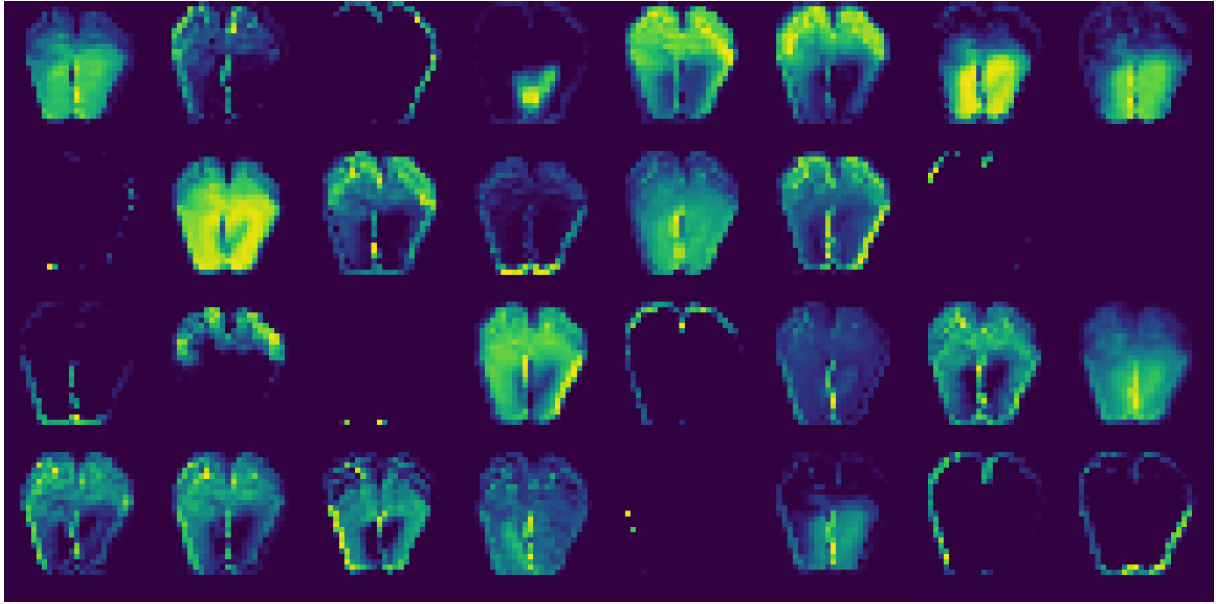

*Figure 1: Visualisation of the latent space representations for a sample feet-segmented thermography image. Conceptually, the 32 low-dimensional (28 x 28 x 1) images shown can be thought of as each viewing the original thermographic high-dimensional (224 x 224 x 3) image through a different series of image filters (convolutional operations). The ConvAE model has optimised these filters such that the original image can be successfully reconstructed by up-sampling all the representations. These representations contain statistically useful thermographic image features for downstream analysis.*

We found compelling evidence for a higher prevalence of DFU risk factors among thermographs assigned to cluster 1 compared to cluster 2 (Table 1). This is particularly evident with measures of the two underlying casual pathologies: peripheral neuropathy and peripheral arterial disease. mTCNS measurements which reliably capture the clinical progression of peripheral neuropathy[24] were much higher in cluster 1 compared to cluster 2, 6 (2 – 13) vs 3 (1 – 5), $p$ = 3.67 x $10^{-6}$. Similarly, TBI measures which quantify vascular insufficiency in PAD[25] (in diabetes populations) were much lower in cluster 1, 0.83 (0.76 – 0.92) vs 0.96 (0.87 – 0.99) in cluster 2, $p$ = 5.17 x $10^{-7}$. There was also strong evidence for higher numbers of diabetes complications, and higher measures of composite DFU risk prediction scores like PODUS – 2020 (including measures of neuropathy, PAD, and history of ulceration/amputation),[26] SIGN (similar measures as PODUS – 2020 but additionally including visual and physical impairment),[27] and Martins-Mendes (original) scores[26] (with measures of neuropathy, PAD, number of diabetes complications, and physical impairment) in cluster 1 compared to cluster 2. Our findings indicate a convergence between thermographic clusters and variables associated with the DFU risk construct.

Further, we found some evidence for higher minimum foot temperatures and lower foot temperature ranges, in cluster 1 compared to cluster 2, 30.11 ± 2.93 vs 29.29 ± 2.76, $p$ = 0.02, and 3.54 ± 1.36 vs 3.99 ± 1.55, $p$ = 0.01, respectively. These results support the trustworthiness of the ConvAE clustering, as they indicate that the clusters are indeed capturing the differences in quantitative foot temperature measurements. As a sensitivity analysis, to show that given a thermograph one can reliably predict cluster assignment, we used supervised- learning to train CNN models and found very high performance on the hold-out test set (AUCs > 0.94).

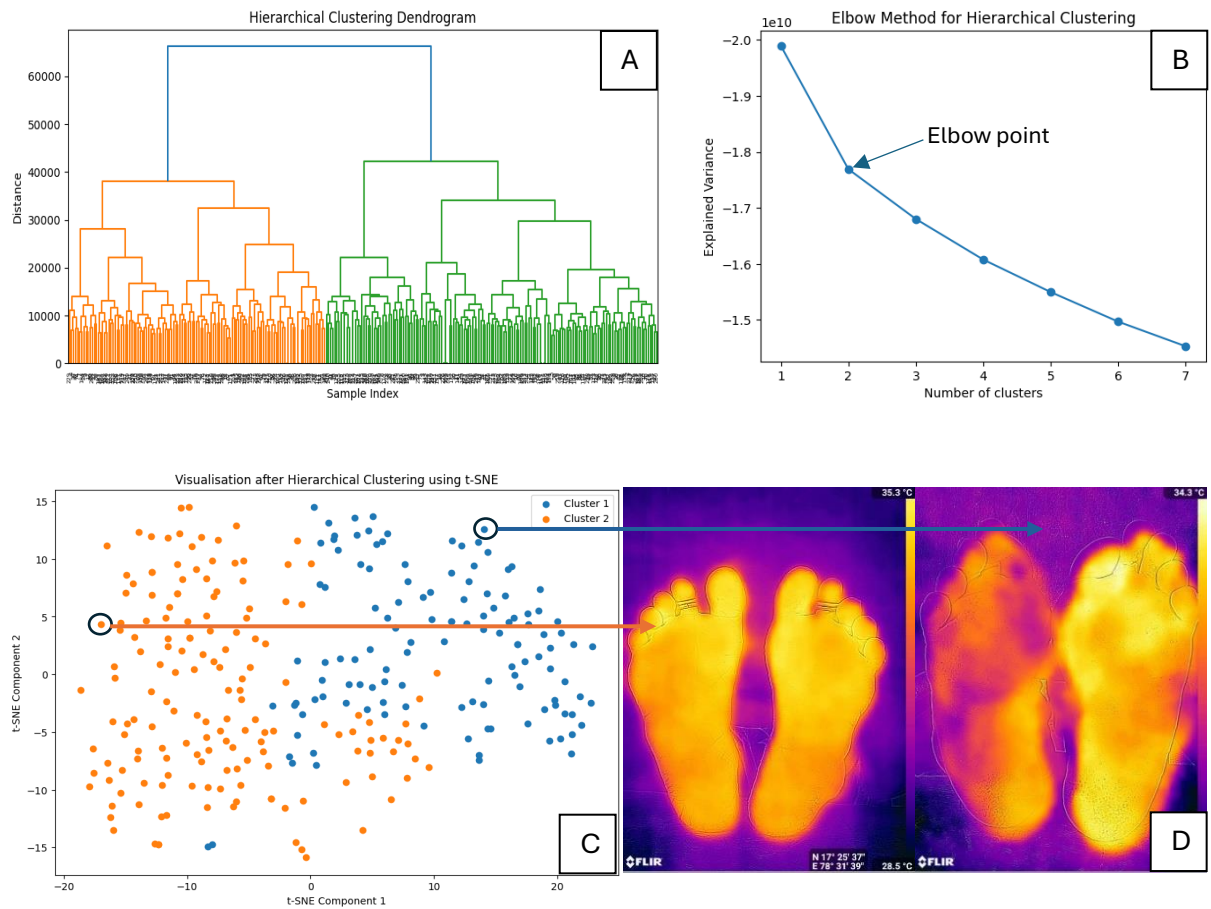

*Figure 2: Visualisation of hierarchical clustering. Panel A – The dendrogram displays the hierarchical relationships among the thermographic images, with the vertical axis representing the distance or dissimilarity between clusters. Each point on the horizontal axis represents each unique thermograph in the dataset. Each of the two primary clusters appears to be subdivided into several sub-clusters. Panel B – Elbow method for determining the optimal number of clusters (denoted by "elbow" point) in hierarchical clustering. The plot displays the explained variance as a function of the number of clusters. Panel C – The t-SNE plot illustrates the two conspicuously differentiated primary clusters with the presence of overlapping data points between the clusters suggesting shared characteristics and a degree of ambiguity in their segregation. Panel D – shows examples of thermographic images from each of the two identified clusters. There is a noticeable thermal asymmetry in the image belonging to cluster 1.*

**Table 1:** *Associations between the intrinsic clusters of thermographic images and patient characteristics*

| Patient Characteristics | Diabetes Patients (N = 282) | | p value |
|---|---|---|---|
| | **Cluster 1: Likely Higher Risk** (*N* = 123) | **Cluster 2: Likely lower Risk** (*N* = 159) | |
| **Demographics** | | | |
| **Age** (years) | 57.50 (51.25 – 62) | 57(52 – 62) | 0.55 |
| **Sex** (Male) | 62 (50.41) | 83 (52.20) | 0.86 |
| **Comorbidity** | | | |
| Current or Former Smoker | 24 (19.51) | 30 (18.87) | 1.00 |
| Alcohol Use Disorders Identification Test Score | 1 (0 – 3) | 1 (0 – 3) | 0.81 |
| Obesity: Body Mass Index ($kg/m^2$) | 25.60 (22.86 – 28.60) | 24.70 (22.29 – 28.18) | 0.29 |
| Atherosclerosis: Carotid intima-media thickness (mm) | 0.62 (0.55 – 0.73) | 0.63 (0.56 – 0.71) | 0.97 |
| Hypertension | 63 (51.22) | 74 (46.54) | 0.51 |
| **Diabetes History** | | | |
| Duration (years) | 6 (3 – 10) | 5 (3 – 9.50) | 0.29 |
| Poor glycaemic control | 53 (43.09) | 86 (54.08) | 0.09 |
| **Peripheral Neuropathy** | | | |
| Light touch – Impaired | 33 (26.83) | 29 (18.24) | 0.11 |
| Position sense – Impaired | 3 (2.44) | 7 (4.40) | 0.58 |
| Pinprick test – Impaired | 25 (20.33) | 13 (8.18) | $5.32 \times 10^{-3}$ |
| Vibration sense – Impaired | 30 (24.39) | 26 (16.35) | 0.13 |
| Temperature sense – Impaired | 35 (28.46) | 45 (28.30) | 1.00 |
| Modified Toronto Clinical Neuropathy Score (mTCNS) | 6 (2 – 13) | 3 (1 – 5) | $3.67 \times 10^{-6}$ |
| Neuropathy (as per mTCNS ≥ 3) | 89 (72.36) | 83 (52.20) | $9.06 \times 10^{-4}$ |
| **Peripheral Arterial Disease (PAD)** | | | |
| Clinical history suggestive of PAD | 34 (27.64) | 26 (16.35) | 0.03 |
| Toe Brachial Index (TBI) | 0.83 (0.76 – 0.92) | 0.96 (0.87 – 0.99) | $5.17 \times 10^{-7}$ |
| PAD (as per TBI ≤ 0.71) | 5 (4.07) | 0 (0.00) | 0.03 |
| **Cardiovascular Disease** | 9 (7.32) | 11 (6.92) | 1.00 |
| **Stroke** | 2 (1.63) | 5 (3.14) | 0.63 |
| **Diabetic Renal Disease** | 1 (0.81) | 2 (1.26) | 1.00 |
| **Diabetic Eye Disease** | 1 (0.81) | 0 (0.00) | 0.52 |
| **Number of Diabetes-Related Complications** | 1 (1 – 2) | 1 (0 – 1) | $2.10 \times 10^{-3}$ |
| **Physical Impairment** | 88 (71.54) | 102 (64.19) | 0.24 |
| **Visual Impairment** | 68 (55.28) | 87 (54.72) | 1.00 |
| **History of Diabetic Foot** | | | |
| Previous history of foot ulcer | 5 (4.07) | 5 (3.14) | 0.93 |
| Previous history of amputations | 0 (0.00) | 1 (0.63) | 1.00 |
| **Composite DFU Risk Scores** | | | |
| PODUS – 2020 (0 – 4) | 1 (1, 1) | 1 (0, 1) | $1.34 \times 10^{-4}$ |
| PODUS – 2020, score ≥ 1 | 95 (77.24) | 89 (55.97) | $3.28 \times 10^{-4}$ |
| SIGN indicative of high-risk foot | 96 (78.05) | 101 (63.52) | 0.01 |
| Martins-Mendes (original) | - 2.47 (-2.74 to -2.14) | -2.74 (-3.29 to -2.47) | $2.62 \times 10^{-3}$ |

**What DFU risk factors can you predict with a thermographic image?**

Performances of fine-tuned CNNs reported on the hold-out testing set revealed poor model performances in both classification (several AUCs merely above 0.50 demonstrating little benefit over random classifiers) and regression (large mean absolute errors that render any subsequent threshold-based clinical classification futile) tasks. Based on poor model performances in predicting measures of neuropathy (regressing mTCNS, or classifying based on mTCNS ≥ 3), or PAD (regressing TBI, or classifying based on clinical history) from thermographs (with or without visual image supplementation), or any of the composite risk scores there is insufficient evidence to support the use of plantar thermography for predicting DFU risk factors in populations with majority previously never-ulcerated diabetes patients.

**Discussion**

There is growing interest in exploring the use of plantar thermography as a digital biomarker for various prediction tasks across the spectrum of diabetic foot disease,[4,28,29] including, classifying foot temperature distributions, [30–33] patterns,[34,35] or contralateral asymmetry among diabetes patient groups,[5,6] detection of visibly smaller DFUs,[36] predicting the risk of DFU,[23,37] monitoring severity and treatment of infections in DFU,[38] and monitoring for DFU recurrence.[9] Several others report advances in technical challenges in thermal image acquisition,[39] registration and/or segmentation,[18,40] automated temperature difference calculations,[6] and user-testing studies[41] all of which are pertinent to establishing thermography as a useful digital biomarker. In this study, we aimed to study the validity of plantar thermography as a digital biomarker to identify patients at high risk of DFU.

**Comparison with previous work**

Our analysis of cross-sectional data showed strong evidence for associations between intrinsic thermographic clusters and several DFU risk factors including casually associated pathologies peripheral neuropathy and PAD, and previously validated composite DFU risk prediction scores PODUS – 2020, Martins-Mendes and SIGN. Our findings are in line with the few previous studies (not performing hot/cold stimulus exposure to limbs) that have found some evidence for associations between thermographic features and measures of neuropathy,[42] or PAD.[11,43,44] Thus, our findings support the construct validity of using thermography for characterising DFU risk.

Interestingly, none of the associations we examined predicted patients at high risk of DFU. There could probably be two reasons for this: (1) spectrum bias[45] owing to a community setting with a relatively lower risk DFU population with short diabetes history, few PAD (TBI ≤ 0.71) and a majority of never-ulcerated patients. Thus the whole spectrum of plantar temperature changes that occur with progressive neuropathy and/or vascular compromise,[46] may not have been represented in our population; (2) thermography and classical risk factors could be characterising incompletely overlapping portions of the DFU risk construct. Theoretically, these measures may relate to DFU risks at varying time frames. Peripheral neuropathy and PAD, both chronic pathologies, can individually or collectively contribute to the risk of DFU, resulting in absolute risk estimates of 3.5% to 21% reported at 2-year time frames.[3] It is plausible that plantar thermography may measure risk across a different or shorter time frame. Very few previous observational studies have developed prediction models specifically for DFU risk using thermographic images. One recent study utilized CNN-based deep learning to classify DFU risk.[37] However, their high-risk target class included diabetes participants with active ulceration, which likely drove the very high classification performance (accuracy: 99.4% for thermal + visual, 97.6% for visual only on a test set of 31 participants).[37] This performance is probably due to the CNN models' ability to distinguish participants with and without visible ulcers, a capability that has been well documented.[47] Another study demonstrated strong classification performance (accuracy: 92.64%) in predicting PAD risk factor status using thermographic imaging, in a population including 23 PAD patients (with ABI < 0.9).[11] However, their images included lower-leg thermographs in addition to plantar thermography.[11] One study used an unsupervised thermographic clustering (k = 3) approach to first subjectively assign each cluster to mild, moderate, or severe DFU risk based on physician review of the thermographs, without detailing the criteria.[23] The accuracy of their models predicting cluster assignments (95.08%) was comparable to our sensitivity analysis results on the same. This is expected, as the ground truth labels used are known to be mathematical functions of the input thermography image. Likewise, models classifying thermographs into thermal change index (a mathematical function of contralateral feet asymmetry adjusted for healthy control reference ranges) classes have also demonstrated excellent performances.[30–33] However, none have previously explored the associations between such possible ground truth labels and variables related to the clinical DFU risk construct. We described strong associations between thermography clusters and DFU risk factors although explored associations did not prove to have predictive value. Taken together, these findings highlight the challenges in identifying appropriate (strongly predictive + clinically relevant) ground truth labels while defining novel digital biomarkers.

**Strengths and Limitations:**

Ours was the largest thermographic study till date to demonstrate thermographic associations with DFU risk factors in the relatively lower-risk group of diabetes patients without previous ulcer histories. We are also among very few studies to successfully demonstrate the feasibility of diabetes foot thermographic imaging using smartphones[4] and a low-cost setup (no use of black insulated material to block out background temperatures) – demonstrating the potential for using thermographic imaging pipelines for diabetes screening in rural/remote lower-middle-income country settings.

**Implications**

The findings from our study justify the need for large-scale prospective studies to (1) study absolute DFU risks across intrinsic thermographic clusters; (2) inform ideal (in terms of predictive efficacy and logistical/economic considerations) thermographic screening interval to identify participants at high risk of DFU; (3) delineate thermographic image patterns and absolute DFU risks across a wider disease spectrum including both previously-ulcerated and never-ulcerated populations.

**Conclusion**

The strong associations between intrinsic thermography clusters and several DFU risk factors support the validity of using thermography for characterising DFU risk. Obtained associations did not prove predictive, likely due to, spectrum bias,[45] or because thermography and classical risk factors characterise incompletely overlapping portions of the DFU risk construct. Our findings highlight the challenges in standardising ground truths when defining novel digital biomarkers.